\documentclass[conference]{IEEEtran}
\IEEEoverridecommandlockouts
\usepackage{cite}
\usepackage{amsmath,amssymb,amsfonts}
\usepackage{algorithmic}
\usepackage{graphicx}
\usepackage{textcomp}
\usepackage{xcolor}

\usepackage{booktabs,threeparttable}
\usepackage{adjustbox}
\usepackage{color,colortbl}
\usepackage{float}
\usepackage{multirow}
\usepackage{array}
\usepackage{lipsum,setspace}

\usepackage{romannum}
\def\BibTeX{{\rm B\kern-.05em{\sc i\kern-.025em b}\kern-.08em
    T\kern-.1667em\lower.7ex\hbox{E}\kern-.125emX}}
\begin{document}

\title{Few-Shot Object Detection via Knowledge Transfer\\
\thanks{This work was supported by Institute for Information \& communications Technology Planning \& Evaluation(IITP) grant funded by the Korea government(MSIT) (No. 2017-0-01779, A machine learning and statistical inference framework for explainable artificial intelligence; No. 2019-0-01371, Development of brain-inspired AI with human-like intelligence; No. 2019-0-00079, Artificial Intelligence Graduate School Program(Korea University)).}
}

\author{\IEEEauthorblockN{Geonuk Kim}
\IEEEauthorblockA{\textit{Dept. Brain and Cognitive Engineering}\\
\textit{Korea University} \\
Seoul, Republic of Korea \\
geonuk\_kim@korea.ac.kr}\\

\and 

\IEEEauthorblockN{Hong-Gyu Jung}
\IEEEauthorblockA{\textit{Dept. Brain and Cognitive Engineering}\\
\textit{Korea University} \\
Seoul, Republic of Korea \\
hkjung00@korea.ac.kr}

\and

\IEEEauthorblockN{Seong-Whan Lee}
\IEEEauthorblockA{\textit{Dept. Artificial Intelligence}\\
\IEEEauthorblockA{\textit{Dept. Brain and Cognitive Engineering}}
\textit{Korea University} \\
Seoul, Republic of Korea \\
sw.lee@korea.ac.kr}
}

\maketitle

\begin{abstract}
 Conventional methods for object detection usually require substantial amounts of training data and annotated bounding boxes. If there are only a few training data and annotations, the object detectors easily overfit and fail to generalize. It exposes the practical weakness of the object detectors. On the other hand, human can easily master new reasoning rules with only a few demonstrations using previously learned knowledge. In this paper, we introduce a few-shot object detection via knowledge transfer, which aims to detect objects from a few training examples. Central to our method is prototypical knowledge transfer with an attached meta-learner. The meta-learner takes support set images that include the few examples of the novel categories and base categories, and predicts prototypes that represent each category as a vector. Then, the prototypes reweight each RoI (Region-of-Interest) feature vector from a query image to remodels R-CNN predictor heads. To facilitate the remodeling process, we predict the prototypes under a graph structure, which propagates information of the correlated base categories to the novel categories with explicit guidance of prior knowledge that represents correlations among categories. Extensive experiments on the PASCAL VOC dataset verifies the effectiveness of the proposed method. 
\end{abstract}


\section{Introduction}
Learning to localize and classify objects in an image is a fundamental problem in computer vision. It has a wide range of applications\cite{roh2010view,maeng2012nighttime}, including robotics, autonomous vehicles and video surveillance. With the success of convolutional neural networks (CNN), great leaps have been achieved in the object detection by remarkable works including Faster-R-CNN\cite{ren2015faster}, Mask R-CNN\cite{he2017mask}, YOLO\cite{redmon2016you} and SSD\cite{liu2016ssd}. Despite of the achievements, most object detectors suffer from an important limitation: they rely on huge amounts of training data and heavily annotated labels. For object detection, annotating the data is very expensive, as it requires not only identifying the categorical labels for all objects in the image but also providing accurate localization information through bounding box coordinates. This warrants a demand for effective object detectors that can generalize well from small amounts of annotated data. Recently, several approaches\cite{kang2019few,yan2019meta} have attempted at resolving the \textit{few-shot object detection} that aims to detect the data-scarce novel categories as well as data-sufficient base categories. Those methods attach a meta-learner to an existing object detector. The meta-learner takes support set images that include the few examples of the novel categories and a subset of examples from the base categories. Given the support images, the meta-learner is expected to predict categorical prototypes which are used to reweight the feature maps from a query image to build category-discriminative feature maps that remodel a prediction layer.
However, in those methods remodelling the prediction layers suffers from a poor embedding space of the prototypes as they predict each prototype independently without considering each others.

In contrast, human vision system is excellent at recognizing new concepts even with less supervision by exploiting prior knowledge accumulated from previous experiences as well as maintaining performance to recognize base concepts\cite{bulthoff2003biologically}. With inspiration from the human vision system, we introduce a few-shot object detection via knowledge transfer (FSOD-KT).

In this theme, we integrate category correlations into a deep neural network to guide exploiting information of base categories to learn the novel concept. To achieve the goal, we construct a \textit{meta-graph} where each node refers to the prototype of a category and each edge represents the semantic correlation between the two corresponding categories, i.e., bicycle is similar with motorbike, and dissimilar with bird. At this point, we define the prior knowledge with a cosine similarity of two corresponding word vectors from category names. With a constructed meta-graph, we prepare to use a graph propagation mechanism to transfer node messages through the graph. In this way, it allows each category to derive information of the correlated categories, and prototypes of novel categories to be predicted with the additional information.

However, simply applying prior knowledge to guide the model is not suitable, as there is a discrepancy between prior knowledge from textual corpus and a given training data distribution\cite{peng2019cm}. To solve the problem, we propagate information along a skip connection as well as a meta-graph to incorporate prior knowledge with the model elastically. Then, with the predicted prototypes, each RoI feature map is reweighted to remodel R-CNN predictor heads to detect novel and base categories in a category-wise reweighted feature map. 

We summarize our contributions as follows: \textbf{1)} We propose to exploit prior knowledge as a form of a graph to guide propagating information to help learn the prototypes of novel categories for few-shot object detection. \textbf{2)} Unlike previous works that merely focus on the novel categories, we aim to detect novel objects while simultaneously maintaining the performance on base categories via considering each other. \textbf{3)} We conduct experiments on PASCAL VOC which is a widely used dataset for object detection and show that the proposed method outperforms existing methods.

\section{Related Works}

\subsection{Object detection}
There are wide range of research areas in computer vision\cite{roh2007accurate,park2013face}. Among them, object detection is a fundamental problem in computer vision. Deep learning based object detection systems are divided into two approaches: a two-stage object detector and a one-stage object detector. R-CNN series\cite{ren2015faster,he2017mask} represent a two-stage object detector to propose RoI (Region-of-Interest) and clarify the RoI with classification and localization. Meanwhile, a one-stage object detector including YOLO\cite{redmon2016you}, SSD\cite{liu2016ssd} and the variants directly detect an object with a single fully convolutional network and show less computation complexity than a two-stage detector as these are proposal free methods. 

\subsection{Few-shot learning} 
Few-shot learning aims to recognize novel objects with a given few training data and corresponding labels. Until now, few-shot learning has achieved a lot of progress. Metric-learning approaches \cite{koch2015siamese,sung2018learning} try to learn an appropriate feature embedding space in which features of same-category examples are similar while those between different categories are dissimilar. Meta-learning approaches \cite{vinyals2016matching,snell2017prototypical,gidaris2018dynamic,finn2017model} design to learn a meta-learner to parameterize the optimization algorithm or predict the parameters of a classifier, so-called \textit{“learning-to-learn”} that can help the model adapt to new environments quickly. Meanwhile, those existing methods simply focus on image recognition. Recently, there are several attempts on few-shot object detection using meta-learning. FSRW\cite{kang2019few} and Meta R-CNN\cite{yan2019meta} apply feature reweighting schemes to a one-stage object detector (YOLOv2\cite{redmon2017yolo9000}) and a two-stage object detector (Faster R-CNN\cite{ren2015faster}), with the help of a meta-learner that takes support images as well as bounding box annotations as inputs. However, those methods suffer base category performance drop during a fine-tuning phase as optimization in predicting prototypes of the novel category adapts the representation in ways that harm the base category. Consequently, it makes an overall performance drop in the methods. 

\subsection{Knowledge transfer visual reasoning}
A wide range of prior knowledge has been explored to incorporate the knowledge with various visual reasoning tasks ranging from image classification to visual relationship reasoning. 
 As a pioneering work, \cite{marino2016more} builds a knowledge graph to correlate object categories and learn graph representation to enhance image representation learning. \cite{wang2018zero} applies graph convolutional network (GCN)\cite{kipf2016semi} to explore semantic interaction and direct map semantic vectors to classifier parameters. To explore high-level tasks, \cite{chen2019knowledge} considers the correlations between specific object pairs and their corresponding relationships to regularize scene graph generation and thus alleviate the effect of the uneven distribution issue. In a few-shot learning scenario, the latest work \cite{li2019large} constructs a categorical hierarchy by semantic clusters and regularizes prediction at each hierarchy. In a zero-shot learning scenario, \cite{changpinyo2016synthesized,changpinyo2020classifier} adopt object attributes as a proxy to learn a visual representation of unseen categories. \cite{frome2013devise,xian2018zero} design to learn semantic embeddings from a text corpus and exploit them to bridge seen and unseen categories. \cite{changpinyo2016synthesized,changpinyo2020classifier} consider combining word embeddings and attributes to predict classifier weights of unseen categories by taking linear combinations of synthetic base classifier weights. Recent works\cite{kampffmeyer2019rethinking,wang2018zero} apply GCN\cite{kipf2016semi} over a WordNet knowledge graph to propagate classifier weights from seen categories to unseen categories. However, none of those works apply knowledge transfer for the few-shot object detection task which is a challenging problem in computer vision.

 \begin{figure*}
\begin{center}
\includegraphics[height=6cm]{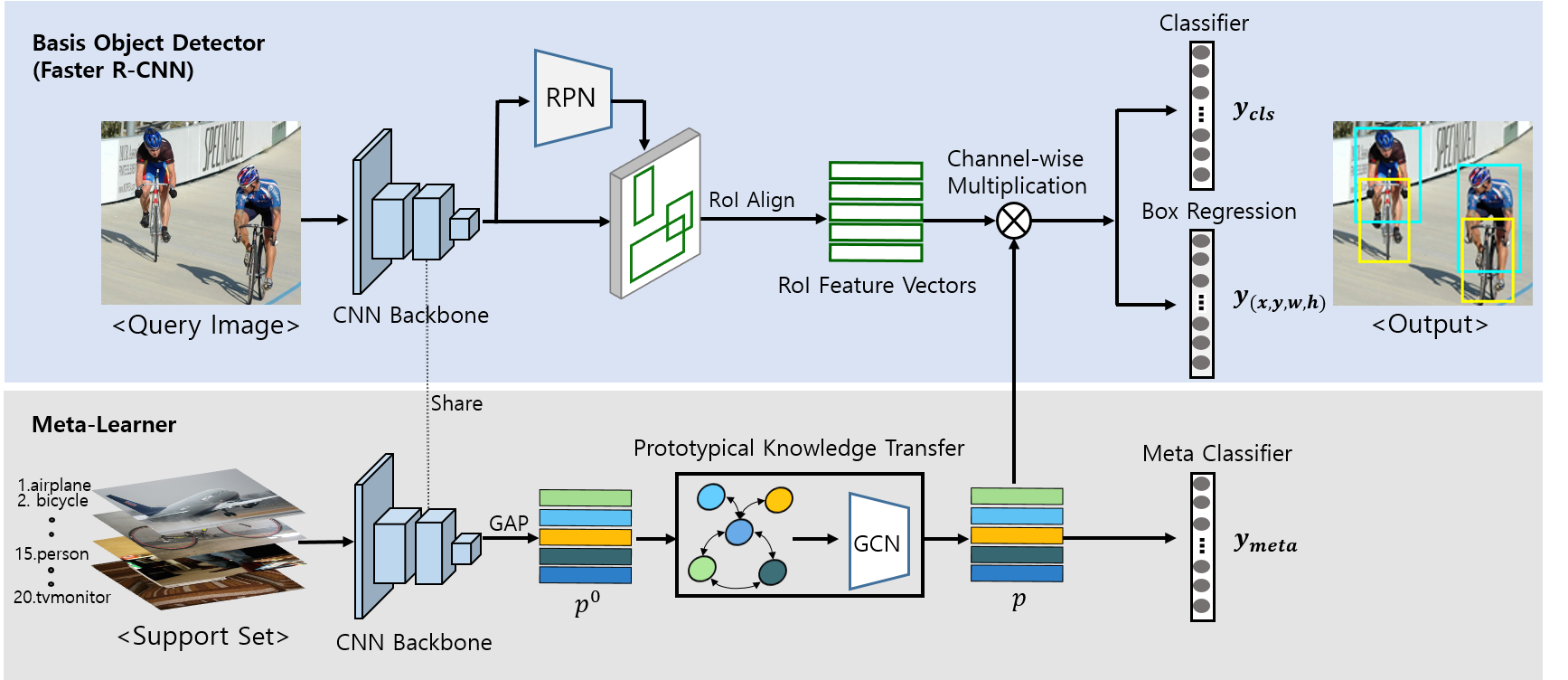}
\end{center}
\caption{An overview of the proposed FSOD-KT. It consists of a basis object detector and a meta-learner. The basis object detector follows a structure of Faster R-CNN to classify objects and regress the corresponding locations. The meta-learner exploits a prior knowledge of semantic correlations between categories to predict categorical prototypes and reweights RoI features.}
\end{figure*}

 \section{Proposed method}
\subsection{Overview}
An overview of the proposed FSOD-KT can be found in Fig. 1. We attach a meta-learner to Faster R-CNN that shares a backbone with Faster R-CNN. The meta-learner aims to predict categorical prototypes from a support set with a prototypical knowledge transfer module that propagates semantic information in a meta-graph structure with prior knowledge. It propagates information along with all pairs in the meta-graph and provides additional information to predict a prototype of each category. Then, the prototypes reweight each RoI feature map extracted from a query image to remodel R-CNN predictor heads to detect objects that are consistent with the prototype represents. 
\subsection{Basis object detector}
We use Faster R-CNN as a basis object detector that is trained on base categories. It has two stages to classify and localize each object in an image. In the first stage, the region proposal network (RPN) predicts each RoI where an object might exist in a feature map from a CNN backbone. Then, RoI feature vectors are sampled with RoI-Align\cite{he2017mask}. At this point, we find that the RPN captures category agnostic objectness features in an image. So, we use the RPN to propose RoIs which are to be classified and localized for both base and novel categories.

 \begin{figure}
\begin{center}
\includegraphics[width=1.0\linewidth,height=3.5cm]{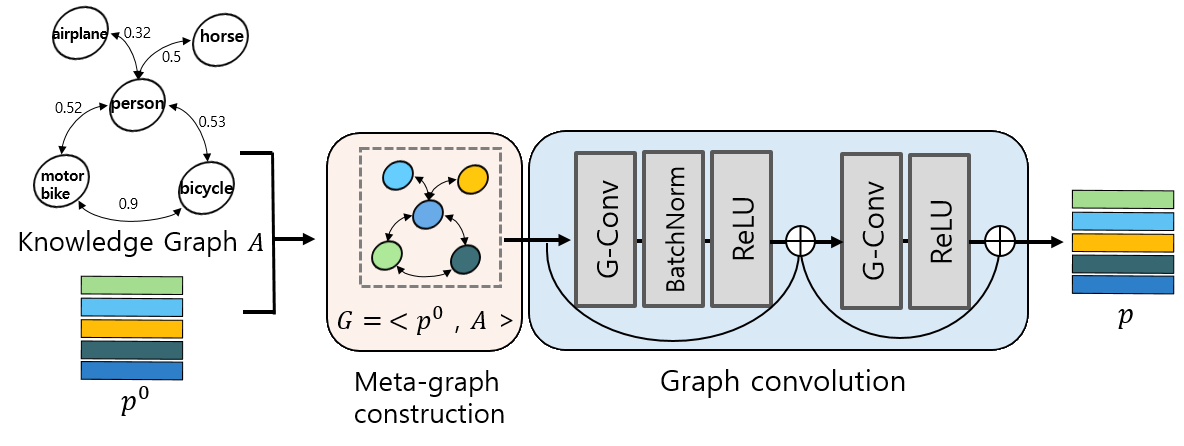}
\end{center}
\caption{The prototypical knowledge transfer module. Prototypical features are aggregated with their neighborhoods through two graph convolution layers with a meta-graph. }

\end{figure}

\subsection{Prototypical knowledge transfer}
The meta-learner $f(\cdot;\theta)$ takes support set images $\{{x_i}\}_{i=1}^{C}$ and predicts prototypes $\{{p_i}\}_{i=1}^{C}$ that represent each category as a vector to remodel the R-CNN predictor heads, where $\theta$ and C denote the parameters of meta-learner and the number of categories, \textit{i.e.}, $C_{base}$ for base a training phase and $C_{base} \cup C_{novel}$ for a fine-tuning phase. To facilitate remodeling R-CNN predictor heads, we introduce a prototypical knowledge transfer module that aims to predict the prototypes with considering each categorical semantic correlation using a graph convolutional network. As shown in Fig. 2, none of the prototypes are predicted independently in this theme. To transfer prototypical knowledge in the graph structure, we first construct a \textit{meta-graph} with initial node \textit{$p^0$} and edge \textit{A} for \textit{C} categories. Preliminary prototype $\textit{$p^0$} \in \mathbb{R}^{\textit{C}\times\textit{S}}$ is defined with global average pooled output features of the meta-learner backbone along with each category, where each node represents one distinct categorical prototype and \textit{S} denotes preliminary prototype features per node. $\textit{A} \in \mathbb{R^{\textit{C}\times\textit{C}}}$ represents semantic correlations among categories in a knowledge graph. These are encoded in the form of a symmetric adjacency matrix. With the constructed meta-graph, we aggregate all node features with each neighborhood node using GCN\cite{kipf2016semi}.
\begin{equation}
H^{(\textit{l}+1)} = \sigma (\textit{D}^{-1}\textit{A}H^{(\textit{l})}\Theta^{(\textit{l})}) 
\end{equation}
More specifically, to predict prototypes \textit{$p$}, we employ a simple propagation rule to perform convolution on the meta-graph following Eq. (1), where $\textit{H}^{(\textit{l})}$ represents activations in the $\textit{l}^{\textit{th}}$ layer and $\Theta \in \mathbb{R^{\textit{S}\times\textit{F}}}$ denotes a trainable weight matrix for layer \textit{l} with \textit{F} corresponding to the number of learned filters. For the first layer, $\textit{H}^{(0)} = \textit{$p^0$}
$. $\sigma{(\cdot)}$ denotes a nonlinear activation function, in our case ReLU. \textit{D} is a degree matrix $ \textit{D} \in \mathbb{R^{\textit{C}\times\textit{C}}}$ which normalizes rows in \textit{A} to ensure that the scale of the feature representations is not modified by \textit{A}.
This propagation gives additional information to predict prototypes of novel categories and generalizes each prototype. On the other hand, there might exist noisy information to some categories to get information propagated. To solve the problem, we exploit a GCN with a residual connection and the ReLU activation function as shown in Fig. 2. Furthermore, to prevent independent features diminished along the aggregation process, we use a skip connection from \textit{$p^0$} to a meta-classifier. It makes the prior knowledge guide visual reasoning on training data elastically. It also can be seen as deep supervision to solve the vanishing gradient problem.
\definecolor{Gray}{gray}{0.9}
\begin{table*}[t]
\caption{Few-shot object detection performance (mAP) on PASCAL VOC 2007 test set in novel categories. We compare our FSOD-KT with existing methods under three different splits of novel categories. mAP indicates mean Average Precision over all novel categories and \textbf{BOLD} indicates state-of-the-art (SOTA)}
\centering
\begin{threeparttable}
\small
\begin{tabular}{l|ccccc|ccccc|ccccc}
\hline
                & \multicolumn{5}{c|}{Novel category setup 1}                                      & \multicolumn{5}{c|}{Novel category setup 2}                                      & \multicolumn{5}{c}{Novel category setup 3}                                     \\ \hline
Method/Shot     & 1             & 2             & 3             & 5             & 10            & 1             & 2             & 3             & 5             & 10            & 1             & 2             & 3             & 5             & 10            \\ \hline
FRCN+joint\cite{yan2019meta}      & 2.7           & 3.1           & 4.3           & 11.8          & 29            & 1.9           & 2.6           & 8.1           & 9.9           & 12.6          & 5.2           & 7.5           & 6.4           & 6.4           & 6.4           \\
FRCN+ft\cite{yan2019meta}         & 11.9          & 16.4          & 29            & 36.9          & 36.9          & 5.9           & 8.5           & 23.4          & 29.1          & 28.8          & 5             & 9.6           & 18.1          & 30.8          & 43.4          \\
FRCN+ft-full\cite{yan2019meta}    & 13.8          & 19.6          & 32.8          & 41.5          & 45.6          & 7.9           & 15.3          & 26.2          & 31.6          & 39.1          & 9.8           & 11.3          & 19.1          & 35            & 45.1          \\
FSRW\cite{kang2019few}   & 14.8          & 15.5          & 26.7          & 33.9          & 47.2          & 15.7          & 15.3          & 22.7          & 30.1          & 39.2          & 19.2          & 21.7          & 25.7          & 40.6          & 41.3          \\
Meta R-CNN\cite{yan2019meta}        & 19.9          & 25.5          & 35            & 45.7          & 51.5          & 10.4          & 19.4          & 29.6          & 34.8          & \textbf{45.4} & 14.3          & 18.2          & 27.5          & 41.2          &\textbf{48.1}          \\
\rowcolor{Gray}
FSOD-KT (ours) & \textbf{27.8} & \textbf{41.4} & \textbf{46.2} & \textbf{55.2} & \textbf{56.8} & \textbf{19.8} & \textbf{27.9} & \textbf{38.7} & \textbf{38.9} & 41.5          & \textbf{29.5} & \textbf{30.6} & \textbf{38.6} & \textbf{43.8} & 45.7 \\ \hline
\end{tabular}
\end{threeparttable}

\end{table*}
\begin{table*}[t]
\caption{AP and mAP on PASCAL VOC 2007 test set for novel categories and base categories of the first base/novel split. We evaluate the performance for 3/10-shot novel-category examples with FRCN under ResNet-101. mAP indicates mean Average Precision over all novel and base categories and \textbf{BOLD} indicates the SOTA}
\begin{adjustbox}{width=1\textwidth+0.1mm}
\Huge
\begin{threeparttable}
\begin{tabular}{cl|cccccc|cccccccccccccccc|c}
\hline
                                         & \multicolumn{1}{c|}{}            & \multicolumn{6}{c|}{Novel categories}                                                            & \multicolumn{16}{c|}{Base categories}                                                                                                                                                                                                                            & \multirow{2}{*}{mAP} \\ \cline{1-24}
\multicolumn{1}{c|}{Shot}                & \multicolumn{1}{c|}{Method} & bird          & bus           & cow           & mbike         & sofa          & mean          & aero          & bike          & boat          & bottle        & car           & cat           & chair         & table         & dog           & horse         & person        & plant         & sheep         & train         & tv            & mean          &                      \\ \hline
\multicolumn{1}{c|}{\multirow{6}{*}{3}}  & FRCN+joint\cite{yan2019meta}                       & 13.7          & 0.4           & 6.4           & 0.8           & 0.2           & 4.3           & \textbf{75.9} & \textbf{80.0} & \textbf{65.9} & \textbf{61.3} & \textbf{85.5} & \textbf{86.1} & \textbf{54.1} & \textbf{68.4} & 83.3 & \textbf{79.1}          & \textbf{78.8} & \textbf{43.7} & \textbf{72.8} & \textbf{80.8} & \textbf{74.7} & \textbf{72.7}          & 55.6                 \\
\multicolumn{1}{c|}{}                    & FRCN+ft\cite{yan2019meta}                          & 31.1          & 24.9          & 51.7          & 23.5          & 13.6          & 29.0          & 65.4          & 56.4          & 46.5          & 41.5          & 73.3          & 84.0          & 40.2          & 55.9          & 72.1          & 75.6          & 74.8          & 32.7          & 60.4          & 71.2          & 71.2          & 61.4          & 53.3                 \\
\multicolumn{1}{c|}{}                    & FRCN+ft-full\cite{yan2019meta}                     & 29.1          & 34.1          & 55.9          & 28.6          & 16.1          & 32.8          & 67.4          & 62.0          & 54.3          & 48.5          & 74.0          & 85.8          & 42.2          & 58.1          & 72.0          & 77.8          & 75.8          & 32.3          & 61.0          & 73.7          & 68.6          & 63.6          & 55.9                 \\
\multicolumn{1}{c|}{}                    & FSRW\cite{kang2019few}                    & 26.1          & 19.1          & 40.7          & 20.4          & 27.1 & 26.7          & 73.6          & 73.1          & 56.7          & 41.6          & 76.1          & 78.7          & 42.6          & 66.8          & 72.0          & 77.7          & 68.5          & 42.0          & 57.1          & 74.7          & 70.7          & 64.8          & 55.2                 \\
\multicolumn{1}{c|}{}                    & Meta R-CNN\cite{yan2019meta}                         & 30.1          & 44.6          & 50.8 & 38.8          & 10.7          & 35.0          & 67.6          & 70.5          & 59.8          & 50.0          & 75.7          & 81.4          & 44.9          & 57.7          & 76.3          & 74.9          & 76.9          & 34.7          & 58.7          & 74.7          & 67.8          & 64.8          & 57.3                 \\
\rowcolor{Gray} \multicolumn{1}{c|}{}                    & FSOD-KT (ours)                  & \textbf{40.5} & \textbf{50.8} & \textbf{59.1}          & \textbf{45.5} & \textbf{34.9}          & \textbf{46.2} & 75.3 & 75.3          & 62.2          & 58.9          & 82.7          & 83.1          & 49.5          & 64.4          & \textbf{83.7}          & 78.2 & 77.8          & 42.1          & 64.8          & 75.0          & 71.8          & 69.6 & \textbf{63.7}        \\ \hline
\multicolumn{1}{c|}{\multirow{6}{*}{10}} & FRCN+joint\cite{yan2019meta}                       & 14.6          & 20.3          & 19.2          & 24.3          & 2.2           & 16.1          & \textbf{78.1} & \textbf{80.0} & \textbf{65.9}          & \textbf{64.1} & \textbf{86.0} & \textbf{87.1} & \textbf{56.9} & \textbf{69.7} & \textbf{84.1} & 80.0          & \textbf{78.4} & \textbf{44.8} & \textbf{74.6} & \textbf{82.7} & \textbf{74.1} & \textbf{73.8} & 63.6                 \\
\multicolumn{1}{c|}{}                    & FRCN+ft\cite{yan2019meta}                          & 31.3          & 36.5          & 54.1 & 26.5          & 36.2          & 36.9          & 68.4          & 75.2          & 59.2          & 54.8          & 74.1          & 80.8          & 42.8          & 56.0          & 68.9          & 77.8          & 75.5          & 34.7          & 66.1          & 71.2          & 66.2          & 64.8          & 57.8                 \\
\multicolumn{1}{c|}{}                    & FRCN+ft-full\cite{yan2019meta}                     & 40.1          & 47.8          & 45.5          & 47.5          & 47.0          & 45.6          & 65.7          & 69.2          & 52.6          & 46.5          & 74.6          & 73.6          & 40.7          & 55.0          & 69.3          & 73.5          & 73.2          & 33.8          & 56.5          & 69.8          & 65.1          & 61.3          & 57.4                 \\
\multicolumn{1}{c|}{}                    & FSRW\cite{kang2019few}                    & 30.0          & 62.7 & 43.2          & \textbf{60.6} & 39.6          & 47.2          & 65.3          & 73.5          & 54.7          & 39.5          & 75.7          & 81.1          & 35.3          & 62.5          & 72.8          & 78.8          & 68.6          & 41.5          & 59.2          & 76.2          & 69.2          & 63.6          & 59.5                 \\
\multicolumn{1}{c|}{}                    & Meta R-CNN\cite{yan2019meta}                         & \textbf{52.5} & 55.9          & 52.7          & 54.6          & 41.6 & 51.5          & 68.1          & 73.9          & 59.8          & 54.2          & 80.1          & 82.9          & 48.8          & 62.8          & 80.1          & \textbf{81.4} & 77.2          & 37.2          & 65.7          & 75.8          & 70.6          & 67.9          & 63.8                 \\
\rowcolor{Gray} \multicolumn{1}{c|}{}                    & FSOD-KT (ours)                  & 50.2          & \textbf{66.3}          & \textbf{60.0}          & 58.8          & \textbf{48.4}          & \textbf{56.8} & 71.9          & 77.8          & 58.2 & 58.1          & 81.3          & 81.7          & 45.2          & 54.0          & 81.4          & 80.5          & 77.4          & 40.7          & 67.3          & 77.2          & 69.5          & 68.1          & \textbf{65.3}        \\ \hline
\end{tabular}
\end{threeparttable}
\end{adjustbox}
\end{table*}
\subsection{Remodeling R-CNN predictor heads}
With the predicted prototypes from a support set with knowledge transfer, we introduce to remodel R-CNN predictor heads. Specifically, our method reweights each RoI $R$ with a prototype $p_i$ for category \textit{i}. 
\begin{equation}
\textit{$R$}_\textit{i} = {R}\otimes {p_i}, \quad{i \in \{1,2,3,...,C\}}
\end{equation}
In this theme, it remodels R-CNN predictor heads to detect objects that are consistent with the prototype represents. Each $\textit{R}_\textit{i}$ produces a detection result which refers to the category \textit{i}. If the highest classification score is lower than an objectness threshold, the RoI is regarded as a background. Note that each prototype and RoI have a same length. Meanwhile, to avoid the ambiguous prediction after RoI-prototype reweighting, we use meta loss $L_{meta}$ with cross entropy as shown in Eq. (3), where $y_i^0$, $y_i$ and $y_i^*$ denote predictions from $p_0$, $p$ and a ground truth label, respectively.  
In this theme, the meta-learner predicts a category-discriminative prototype while predicting the label $y_i$ corresponding to the support image $x_i$.

\begin{equation}
    L_{meta}=-\sum_{i=1}^{C} y_i^*\frac {logP(y_i|x_i)+logP(y_i^0|x_i)} {2}
\end{equation}
It encourages each prototype to fall in the category each object belongs to. Finally, we define a loss on each sampled RoI as Eq. (4),
\begin{equation}
    L=L_{cls}(\theta,\phi)+L_{box}(\theta,\phi)+L_{meta}(\theta)
\end{equation}
where $\phi$ denotes the parameters of Faster R-CNN. The classification loss $L_{cls}$ and bounding-box loss $L_{box}$ are identical as those defined in \cite{ren2015faster}. 
\section{Implementation details}
\subsection{Mini batch construction}
We follow \cite{kang2019few} to construct a mini-batch. In a base training phase, there exist a large scale training images for base categories. If a training image contains novel categories, we regard the region as background. In a fine-tuning phase, to train with \textit{K}-shot samples along \textit{N} novel categories, we randomly select instance-wise \textit{K}-shot samples along with \textit{N} novel categories. Note that more than one instance can exist in a single image. Therefore, less than \textit{K} images can be selected per category. We construct a mini-batch with images that contain \textit{K} instances for both base and novel categories. The samples are shared along a query image and a support set, and binary masks representing a location of the objects in an image are attached to the support set to exploit spatial information in an image.

\subsection{Knowledge graph construction}
To construct a knowledge graph that represents semantic correlations between each category, we calculate a cosine similarity between each word vector of category names from the pretrained GloVe\cite{pennington2014glove} text model. In this way, two categories show high cosine similarity if their word vectors are embedded with close. More specifically, if the categories $i$ and $j$ represent word embedding $w_i$ and $w_j$, we calculate a cosine similarity between them to define the knowledge graph.
\begin{equation}
\cos ({w_i},{w_j})= \frac{{w_i} \cdot {w_j}^T}{\|{ w_i}\|_2 \cdot \|{w_j}\|_2}
\end{equation}
\subsection{Training details}
We use ResNet-101 as a backbone feature extractor. All models are trained using the SGD optimizer with the batch size of 8, the momentum of 0.9 and the weight decay of 0.001. The learning rate of 0.01 is used during the base training phase and 0.001 during the few-shot fine-tuning phase.

\begin{figure*}
\centering
\begin{center}
\includegraphics[width=1.0\linewidth,height=4.4cm]{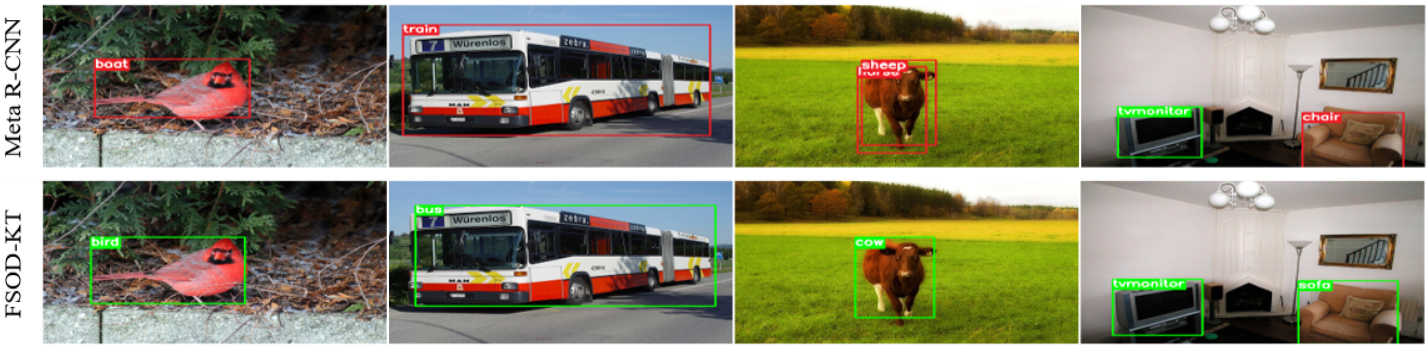}
\end{center}
\caption{Examples of novel category objects \textit{(bird, bus, cow, sofa)} detected by Meta R-CNN and our FSOD-KT. \textcolor{red}{RED} and \textcolor{green}{GREEN} indicate failure and success cases, respectively. Our method is more accurate than the existing method due to the help of the prototypical knowledge transfer.}
\end{figure*}

\begin{table}
\begin{center}
\centering
\caption{Ablation study based on skip connection between preliminary prototypes and a meta classification layer}
\begin{threeparttable}
\small
\begin{tabular}{c|c|c|c}
\hline
Shot                & Method               & Base AP & Novel AP \\ \hline
\multirow{2}{*}{3}  & skip connection (w/o) & 68.3  & 44.3   \\
                    & skip connection (w)   & \textbf{69.6} & \textbf{46.2}  \\ \hline
\multirow{2}{*}{10} & skip connection (w/o) & 67.7  & 52.1   \\
                    & skip connection (w)   & \textbf{68.1} & \textbf{56.8}  \\ \hline
\end{tabular}
\end{threeparttable}
\end{center}
\end{table}

\begin{table}
\begin{center}
\caption{Experimental comparison with randomly generated prior knowledge}
\begin{threeparttable}
\small
\begin{tabular}{c|c|c|c}
\hline
Shot                & Method               & Base AP & Novel AP \\ \hline
\multirow{2}{*}{3}  & Uniform random & 62.4  & 42.6   \\
                    & Correlation value (ours)   & \textbf{69.6} & \textbf{46.2}  \\ \hline
\multirow{2}{*}{10} & Uniform random & 52.2  & 54.8   \\
                    & Correlation value (ours)   & \textbf{68.1} & \textbf{56.8}  \\ \hline
\end{tabular}
\end{threeparttable}
\end{center}
\end{table}

\section{Experimental results}
\subsection{Benchmarks and baselines}
\subsubsection{Benchmark dataset}
Following the previous works\cite{kang2019few,yan2019meta}, we evaluate on PASCAL VOC covering 20 object categories with three kinds of novel set split settings. The novel categories are instance-wise sampled \textit{K} = 1, 2, 3, 5, 10 times from trainval of PASCAL VOC 2007+2012 for training. PASCAL VOC 2007 test set is used to evaluate mAP (mean Average Precision) score and we adopt the PASCAL Challenge protocol that a correct prediction should have more than 0.5 IoU (Intersection over Union) with a ground truth. We evaluate on three different novel/base category split settings: \textbf{Novel-category split 1} (\textit{“bird”, “bus”, “cow”, “mbike”, “sofa”/ rest}); \textbf{Novel-category split 2} (\textit{“aero”, “bottle”,“cow”,“horse”,“sofa” / rest}) and \textbf{Novel-category split 3} (\textit{“boat”, “cat”, “mbike”,“sheep”, “sofa”/ rest}).
\subsubsection{Existing baselines}
The proposed method is compared with \textbf{FRCN+joint}, \textbf{FRCN+ft}, \textbf{FRCN+ft-full}, \textbf{FSRW}\cite{kang2019few} and \textbf{Meta R-CNN}\cite{yan2019meta}. FRCN+joint is to jointly train original Faster R-CNN with abundant base category data and a few novel category data. FRCN+ft is to train the model with two-stage phases but without meta-learner. FRCN+ft-full follows a same strategy with FRCN+ft but fully converged in a fine-tuning phase. We also compare our proposed method with the few-shot object detectors FSRW\cite{kang2019few} and Meta R-CNN\cite{yan2019meta}, and follow the evaluation results of existing methods reported in those methods.

\subsection{Results on PASCAL VOC}
The experimental results are shown in Table \Romannum {1}. The \textit{K}-shot object detection is performed in three different novel/base category splits. It shows a generic object detector with joint training or simply fine-tuning is poor to detect novel categories. The proposed method outperforms the existing methods.

For more detailed comparison, Table \Romannum {2} shows AP and mAP comparison across the categories based on the first novel/base split with \textit{K} = 3, 10. FRCN+joint outperforms other methods on base category performance. But at the same time, it shows extremely poor performance on the novel category that means it is not plausible to use a conventional object detector to learn the novel category with few training examples. For the novel categories, even though there are some cases that existing methods show higher AP on certain categories, FSOD-KT shows higher mAP across all novel categories than existing methods. For base categories, we find that base category performance drop is so marginal that the proposed method outperforms FSRW and Meta R-CNN on the base category as well. Particularly in 3-shot setting, the margin for the novel category and the base category is 19.5\%, 4.8\% and 11.2\%, 4.8\% with FSRW and Meta R-CNN, respectively. As a result, our FSOD-KT shows higher mAP across all categories than existing methods. In Fig. 3, we also visualize qualitative comparisons with Meta R-CNN and our FSOD-KT.

\subsection{Ablation studies}
We conduct ablation studies to clarify how components in our method affect the final performance. The experiments are on PASCAL VOC base/novel split 1 with \textit{K} = 3, 10. 

\subsubsection{Skip connection in prototypical knowledge transfer}
The proposed method applies skip connection between preliminary prototypes and meta classifier to make prior knowledge guide visual reasoning elastically. To figure out the effect of skip connection, we evaluate our method without skip connection and compare it with the proposed model. As Table \Romannum {3} shows, we observe that when the prototypical knowledge is transferred without skip connection, Novel AP degrade about 1.9\% and 4.7\% on 3-shot and 10-shot setting, respectively.

\subsubsection{Knowledge graph as a prior}
To guide informative components to be propagated between categories, we exploit prior knowledge from a cosine similarity among word vectors of category names. To figure out the effect of prior knowledge to guide prototypical knowledge transfer, we replace the prior knowledge with randomly generated values from a uniform distribution. The comparison results are presented in Table \Romannum {4}. As we introduced, when the prior knowledge represents each correlation between categories, the effect is more remarkable than the case randomly generated. Meanwhile, we find that the AP of novel categories is degraded but still outperforms existing methods. One possible reason is that it can still make a chance to predict more generalized prototypes of novel categories with additional information even though the guidance of knowledge is noisy.

\section{Conclusion}
In this paper, we have presented a novel few-shot object detection network that incorporates prior knowledge of category correlations to guide exploiting knowledge from base categories to help learn the prototypes of novel categories. For the purpose, we propose a prototypical knowledge transfer module with a meta-graph representing categorical correlations, and prepare to use a graph convolutional network to aggregate and refine the prototypes. In this way, the predicted prototypes facilitate modeling R-CNN predictor heads to detect objects that are consistent with the prototype represents. We also figure out simply exploiting prior knowledge to guide visual reasoning is not effective as there is a discrepancy between prior knowledge and given data. We introduce to use skip connection in a prototypical knowledge transfer module to solve the problem. Extensive experiments on the widely used PASCAL VOC dataset demonstrate the effectiveness of our FSOD-KT over existing methods.


\bibliographystyle{IEEEtran.bst}
\bibliography{fsod.bib}

\end{document}